\documentclass[runningheads]{llncs}
\usepackage{graphicx}
\usepackage{hyperref}

\usepackage[utf8]{inputenc}
\usepackage{booktabs}
\usepackage{tikz}
\usepackage{multirow}
\usepackage{nicefrac}
\usepackage{amssymb}

\makeatletter
\newcommand{\printfnsymbol}[1]{\textsuperscript{\@fnsymbol{#1}}}
\makeatother

\begin{document}
\title{Active Learning for Entity Alignment}
\author{
Max Berrendorf\inst{1}\thanks{equal contribution}\and
Evgeniy Faerman\inst{1}\printfnsymbol{1} \and
Volker Tresp\inst{1,2}
}

\institute{
Ludwig-Maximilians-Universität München, Munich, Germany\\
\email{\{berrendorf,faerman\}@dbs.ifi.lmu.de}
\and
Siemens AG, Munich, Germany\\
\email{volker.tresp@siemens.com}
}
\authorrunning{Berrendorf, Faerman, Tresp.}

\maketitle              \begin{abstract}
In this work, we propose a novel framework for labeling entity alignments in knowledge graph datasets. Different strategies to select informative instances for the human labeler build the core of our framework.
We illustrate how the labeling of entity alignments is different from assigning class labels to single instances and how these differences affect the labeling efficiency.  Based on these considerations, we propose and evaluate different active and passive learning strategies. One of our main findings is that passive learning approaches, which can be efficiently precomputed, and deployed more easily, achieve performance comparable to the active learning strategies. 
In the spirit of reproducible research, we make our code available at \url{https://github.com/mberr/ea_active_learning}. 
\end{abstract}

\section{Introduction}
A knowledge graph (KG) is a way to store information (semi-)structurally to enable automatic data processing and data interpretation.
KGs are utilized in various Information Retrieval related applications requiring semantic search of information \cite{bast2016semantic,irtutorial}.  
While there exist various large open-source KGs, such as YAGO-3 \cite{DBLP:conf/cidr/MahdisoltaniBS15}, Wikidata \cite{vrandevcic2014wikidata}, or ConceptNet \cite{speer2017conceptnet}, they often contain orthogonal information, and have their respective strength and weaknesses.
Hence, being able to combine information from different knowledge graphs is required in many applications.
An important subtask is identifying matching entities across several graphs, called \emph{entity alignment} (EA).
Recent years witnessed substantial advances regarding the methodology, in particular involving graph neural networks (GNNs)
\cite{cao2019multi,chen2017multigraph,guo2019learning,pei2019semi,sun2017cross,sun2018bootstrapping,sun2019knowledge,trisedya2019entity,wang2018cross,xu2019crosslingual,zhang2019multi,zhu2019neighborhood}.
Common among these approaches is that they use a set of given seed alignments and infer the remaining ones.
While several benchmark datasets are equipped with alignments, acquiring them in practice is cumbersome and expensive, often requiring human annotators.
To address this problem, we propose to use \emph{active learning} for entity alignment.
In summary, our contributions are as follows:
\begin{itemize}
    \item To the best of our knowledge, we are the first to propose using active learning for entity alignment in knowledge graphs. We investigate and formalize the problem, identify critical aspects, and highlight differences to the classical active learning setting for classification.
    \item A specialty of entity alignment is that learning is focused on information about aligned nodes. We show how to additionally utilize information about exclusive nodes in an active learning setting, which leads to significant improvements.
    \item We propose several different heuristics, based upon node centrality, graph and embedding coverage, Bayesian model uncertainty, and certainty matching.
    \item We thoroughly evaluate and discuss the heuristics' empirical performance on a well-established benchmark dataset using a recent GNN-based model. Thereby, we show that state-of-the-art heuristics for classification tasks perform poorly compared to surprisingly simple node centrality based approaches.
\end{itemize} \section{Problem Setting}
We study the problem of entity alignment for knowledge graphs (EA).
A knowledge graph can be represented by the triple $\mathcal{G} = (\mathcal{E}, \mathcal{R}, \mathcal{T})$, where $\mathcal{E}$ is a set of entities, $\mathcal{R}$ a set of relations, and $\mathcal{T} \subseteq \mathcal{E} \times \mathcal{R} \times \mathcal{E}$ a set of triples.
The alignment problem now considers two such graphs $\mathcal{G}^L, \mathcal{G}^R$ and seeks to identify entities common to both, together with their mapping.
The mapping can be defined by the set of matching entity pairs $\mathcal{A} = \{(e, e') \mid e \in \mathcal{E}^L, e' \in \mathcal{E}^R, e \equiv e' \}$, where $\equiv$ denotes the matching relation.
While some works are using additional information such as attributes or entity labels, we solely consider the graph structure's relational information.
Thus, a subset of alignments $\mathcal{A}_{train} \subseteq \mathcal{A}$ is provided, and the task is to infer the remaining alignments $\mathcal{A}_{test} := \mathcal{A} \setminus \mathcal{A}_{train}$.
With $\mathcal{A}^L := \{e \in \mathcal{E}^L \mid \exists e' \in \mathcal{E}^R: (e, e') \in \mathcal{A}\}$ we denote the set of entities from $\mathcal{G}^L$ which do have a match in $\mathcal{A}$, and $\mathcal{A}^R$ analogously.
With $\mathcal{X}^L = \mathcal{E^L} \setminus \mathcal{A}^L$ we denote the set of exclusive entities in the graph $\mathcal{G}^L$ which occur neither in train nor test alignment, and $\mathcal{X}^R$ analogously.

In practice, obtaining high-quality training alignments means employing a human annotator.
As knowledge graphs can become large, annotating a sufficient number of alignment pairs may require significant labeling efforts and might be costly.
Thus, we study strategies to select the most informative alignment labels to achieve higher performance with fewer labels, commonly referred to as active learning.
The following section surveys existing literature about active learning with a particular focus on graphs and reveals differences in our setting. \section{Related Work}
Classical active learning approaches \cite{settles2009active} often do not perform well in batch settings with neural network architectures.
Therefore, developing active learning heuristics for neural networks is an active research area.
New approaches were proposed for image \cite{beluch2018power,gal2017deep,geifman2017deep,sener2017active,wang2016cost,yang2017suggestive}, text \cite{shen2017deep,zhang2017active} and relational \cite{cai2017active,gao2018active,li2019semi,ostapuk2019activelink,wu2019active} data.
Active learning algorithms aim to select the most informative training instances. For instance, the intuition behind uncertainty sampling \cite{lewis1994heterogeneous} is that instances about which the model is unconfident comprise new or not yet explored information.
However, the estimation of neural networks' uncertainty is not a trivial task since neural networks are often overconfident about their predictions \cite{gal2016dropout}.
One approach to tackle this problem is to use Monte-Carlo dropout to estimate the uncertainty for active learning heuristics \cite{gal2017deep,ostapuk2019activelink,shen2017deep}. Alternatively,
\cite{beluch2018power} demonstrated that ensembles of different models lead to better uncertainty estimation and consequently better instance selection. 
The method described in \cite{li2019semi} adopts a different approach and queries labels for instances for which it is the most certain that they are unlabeled.
For this assessment, the authors propose an adversarial framework, where the discriminator differentiates between labeled and unlabeled data.

Geometric or density-based approaches \cite{cai2017active,gao2018active,geifman2017deep,sener2017active,wu2019active,yang2017suggestive}, on the other hand, aim to select the most representative instances. Therefore, unlabeled instances are selected for labeling, such that labeled instances cover unlabeled data in the embedding space.
Other approaches to estimate the informativeness of unlabeled samples use, e.g., the expected length of gradient \cite{zhang2017active}.

Active learning approaches with neural networks on relational data were so far applied to the classification of nodes in homogeneous graphs \cite{cai2017active,gao2018active,li2019semi,wu2019active} and link prediction in knowledge graphs \cite{ostapuk2019activelink}. In \cite{cortes2012active,cortes2013active,malmi2017active} authors propose active learning approaches for the graph matching problem, where the matching costs are known \emph{in advance}, and the goal is to minimize assignment costs. Note that this is different from our task, where the goal is to learn meaningful representations of the entities. \section{Methodology}
In this section, we introduce our proposed labeling setting and describe data post-processing to leverage exclusive nodes.
Moreover, we propose numerous new labeling strategies: Some strategies take inspiration from existing state-of-the-art heuristics for classification. Others are developed entirely new based on our intuitions. Finally, we present our evaluation framework for the evaluation of different heuristics.

\subsection{Labeling Setting}
Since we are dealing with matching KGs, where entities have meaningful labels, we assume that human annotators use these entity names for matching.
Therefore, we see two different possibilities to formulate the labeling task:
\begin{enumerate}
    \item The system presents annotators with possible matching pairs, and they label it as \texttt{True} or \texttt{False}
    \item The system presents annotators a node from one of the two KGs, and the task is to find all matching nodes in the other KG.
\end{enumerate}
It is easier to label a single instance in the first scenario, as it is a yes/no question. However, since each node can have more than one matching node in the other KG, $|\mathcal{E}^L| \times |\mathcal{E}^R|$ queries are necessary to label the whole dataset.
In contrast, in the second scenario, human annotators need a similar qualification but the time spent per labeled instance increases because they have to search for possible matchings. However, there are the following advantages of the second scenario:

First, there are only $|\mathcal{E}^L| + |\mathcal{E}^R|$ possible queries.
Second, in both scenarios, the learning algorithm needs positive matchings to start training. Assuming $|\mathcal{A}^L| \approx |\mathcal{A}^R| \approx |\mathcal{A}|$ and $|\mathcal{E}^R| \approx |\mathcal{E}^L| \approx |\mathcal{E}|$, the probability to select a match with a random query is in the first scenario $\nicefrac{|\mathcal{A}|}{|\mathcal{E}|^2}$, whereas for the second scenario it is $\nicefrac{|\mathcal{A}|}{|\mathcal{E}|}$.
Additionally, in the second scenario, it is possible to start with some simple graph-based heuristics, e.g., based on a graph centrality score like degree or betweenness. For many KGs, it is a valid assumption that the probability of having a match is higher for more central nodes. Cold-start labeling performance is especially relevant when the labeling budget is restricted. 
Third, in the classical active learning scenario, there is the assumption that each query returns a valid label. However, for EA, the information that two nodes do not match is limited since negative examples can also be obtained by negative sampling. In contrast, in the second scenario, we can use information about missing matchings to adapt the dataset, see Section~\ref{:subsection:post_processing}.

In this paper, we focus on the second scenario. 
However, heuristics relying on information from the matching model described in Section~\ref{:subsection:heuristics} can also be applied in the first scenario. 

\subsection{Dataset Adjustment}
\label{:subsection:post_processing}
The EA task's main motivation is either the fusion of knowledge into a single database or exchanging information between different databases.
In both cases, the primary assumption is that there is information in one KG, which is not available in the other.
This information comes in relations between aligned entities, relations with exclusive entities, or relations between exclusive entities.
While larger differences between the KGs increase their fusion value, they also increase the difficulty of matching processes.
One possibility to partially mitigate this problem is to enrich both KGs independently using link prediction and transfer links between aligned entities in the training set \cite{cao2019multi,li2019semi}.
As this methodology does only deal with missing relations between shared entities, in this work, we go a step further:
Since we control the labeling process, we naturally learn about exclusive nodes from the annotators.
Therefore, we propose to remove the exclusive nodes from the KGs for the matching step. After the matching is finished, the exclusive nodes can be re-introduced.
In the classical EA setting, where the KGs and partial alignments are already given, and there is no control over dataset creation, the analogous removal of exclusive nodes is not possible:
To determine whether a node is exclusive or just not contained in the training alignment requires access to the test alignments, hence representing a form of test leakage.

\subsection{Active Learning Heuristics}
\label{:subsection:heuristics}
The main goal of active learning approaches is to select the most informative set of examples. In our setting, each query either results in matches or verified exclusiveness, both providing new information.
Nodes with an aligned node in the other KG contribute to the signal for the supervised training.
State-of-the-art GNN models for EA learn by aggregating the k-hop neighborhood of a node.
Two matching nodes in training become similar when their aggregated neighborhood is similar.
Therefore, the centrality of identified alignments or their coverage is vital for the performance.
On the other hand, exclusive nodes improve training by making both KGs more similar.
Since it is not clear from the outset, what affects the final performance most, we analyze heuristics with different inductive biases.
\\
\textbf{Node Centrality --}
Selecting nodes with high centrality in the graph has the following effects:
(a) a higher probability of a match in the opposite graph, and (b) updates for a larger number of neighbors if a match or significant graph changes when being exclusive.
Although there is a large variety of different centrality measures in graphs \cite{das2018study}, we observed in initial experiments that they perform similarly. Therefore, in this work, we evaluate two heuristics based on the nodes' role in the graph. The first, \emph{degree} heuristic (denoted as \emph{deg}), orders nodes by their degree, and the nodes with a higher degree are selected first. The second, \emph{betweenness} heuristic (\emph{betw}), works similarly and relies on the betweenness centrality measure.
\\
\textbf{Graph Coverage --}
Real-World graphs tend to have densely connected components \cite{faerman2018lasagne}.
In this case, if nodes for labeling are selected according to some centrality measure, there may be a significant overlap of neighborhoods. At the same time, large portions of the graph do receive no or infrequent updates.
Therefore, we propose a heuristic, seeking to distribute labels across the graph.
We adopt an \emph{approximate vertex cover} algorithm \cite{Puthal2015} to define an active learning heuristic for entity alignment.
Each node is initialized with a weight equal to its degree.
Subsequently, we select the node from both graphs with the largest weight, remove it from the candidate list, and decrease all its neighbors' weight by one. We denote this heuristic as \emph{avc}.
\\
\textbf{Embedding Space Coverage --}
The goal of embedding space coverage approaches is to cover the parts of the embedding space containing data as well as possible.
Here we adapt the state-of-the art method \emph{coreset} \cite{sener2017active} (denoted as \emph{cs}) for the EA task.
Thereby, we aim to represent each graph's embedding space by nodes with \emph{positive} matchings.
We adopt a greedy approach from \cite{sener2017active}, which in each step selects the object with the largest distance to the nearest neighbor among already chosen items. Its performance was similar to the mixed-integer program algorithm while being significantly faster.
In the process of node selection, it is not known whether nodes in the same batch have matchings or are exclusive.
Thereby, in each step, each candidate node is associated with a score according to its distance to the nearest positive matching or the nodes already selected as potential positives in the same batch.
The node with the largest distance to the closest positive point is added to the batch.
\\
\textbf{Embedding Space Coverage by Central Nodes --}
\begin{figure}
    \centering
    \includegraphics[width=.31\linewidth]{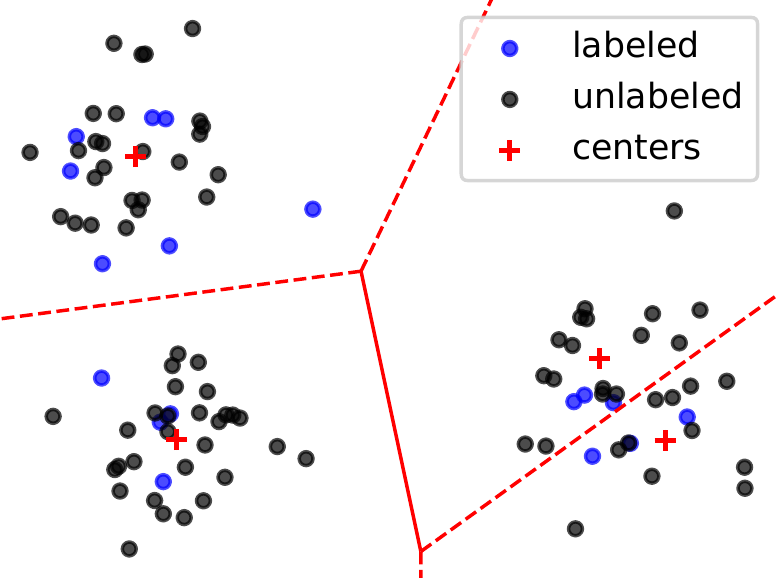}
    \caption{
    Schematic visualization of the \emph{esccn} heuristic.
    The labeled nodes per cluster are counted and used to derive how many samples to draw from this cluster.
    Another heuristic is then used to select the specific number from the given clusters, e.g., a graph-based \emph{degree} heuristic.
    }
    \label{fig:mpiu}
\end{figure}
The possible disadvantage of \emph{coreset} heuristic in the context of entity alignment is that selected nodes may have low centrality and therefore affect only a small portion of the graph. Intuitively, it is possible because each next candidate is maximally distant from all nodes with positive matchings, which are expected to be more or less central. In this heuristic, we try to remedy this effect and sample nodes with high centrality in different parts of embedding space. Therefore, in each step, we perform clustering of node representations from both graphs in the joint space, c.f. Figure~\ref{fig:mpiu}. We count already labeled nodes in each cluster and determine the number of candidates selected from this specific cluster. This number is inversely proportional to the number of already labeled nodes in the cluster. We then use a node centrality based heuristic to select the chosen number of candidates per cluster.
We denote this heuristic by \emph{esccn}.
\\
\textbf{Uncertainty Matching --}
Uncertainty-based approaches are motivated by the idea that the most informative nodes are those for which the model is most uncertain about the final prediction.
We reformulate EA as a classification problem: The number of classes corresponds to the number of matching candidates, and we normalize the vector of similarities to the matching candidates with the softmax operation. 
A typical uncertainty metric for classification is \emph{Shannon entropy} computed over the class probability distribution, where large entropy corresponds to high uncertainty.
We can employ \emph{Monte-Carlo Dropout} to compute a Bayesian approximation of the softmax for the entropy similarly to \cite{gao2018active}. However, the repeatable high entropy across multiple dropout masks indicates the \emph{prediction uncertainty}, where the model is \emph{certain} that a right prediction is impossible.
In the context of entity alignment, we expect high prediction uncertainty for the exclusive nodes since a model may be \emph{certain} about lacking good matchings. Therefore we opt for model uncertainty for the entity alignment. The model uncertainty is high if the model makes different (certain) decisions for the same instances in multiple runs \cite{gal2016uncertainty}. We employ \emph{BALD} \cite{houlsby2011bayesian} with Monte-Carlo Dropout \cite{gao2018active}. The heuristic computes the expected difference between the entropy of single model prediction and expected entropy.
Note that numerous classes may lead to similar entropy and BALD values for the whole dataset.
To mitigate this effect, we employ softmax temperature \cite{hinton2015distilling}.
\\
\textbf{Certainty Matching --}
\begin{figure}
    \centering
    \includegraphics[width=.4\linewidth]{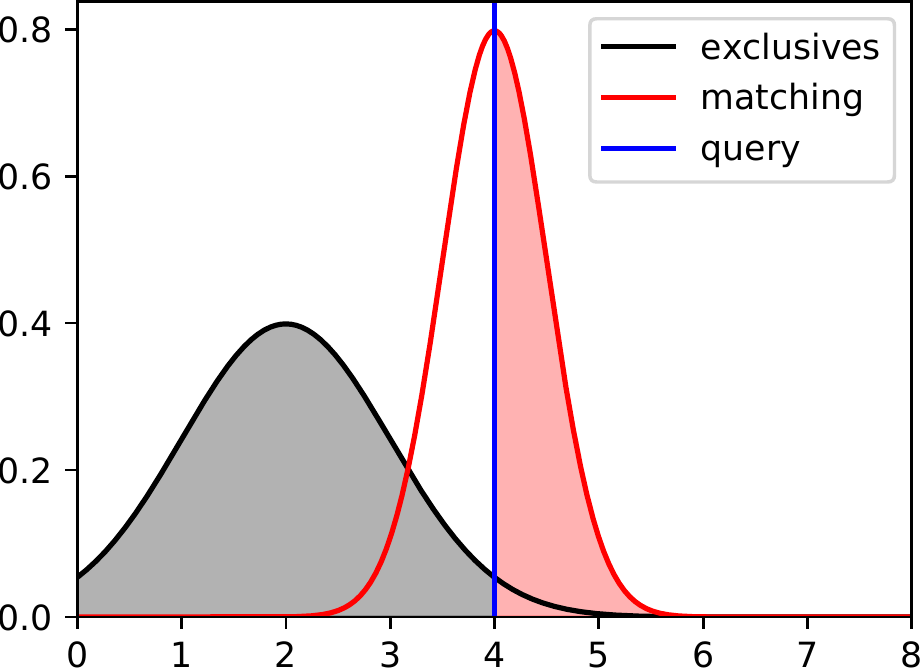}
    \caption{
    Visualization of scoring method of the \emph{prexp} heuristic.
    We fit two normal distributions for matching and exclusive nodes.
    Each distribution models the maximum similarity these nodes have to any node in the other graph ($s_{max}(q)$).
    To assess the quality of a query $q$, we get its maximum similarity, and evaluate $P_{match}(s_{max}(e)\leq s_{max}(q)) -P_{excl}(s_{max}(e) \geq s_{max}(q))$, i.e. the black area minus the red one.
    }
    \label{fig:peb}
\end{figure}
A distinctive property of EA is that the supervised learning signal is provided only by the part of the labeled nodes that have a matching partner in the other graph.
Therefore, we propose a heuristic that prefers nodes having matches in the opposite graph, named \emph{previous-experience-based (prexp)}.
As the model is trained to have high similarities between matching nodes, the node with maximum similarity is the most likely matching partner for a given node.
Moreover, we expect that higher similarity values indicate a better match, such that we can utilise this maximum similarity as a matching score: $s_{max}(e)=\max_{e' \in \mathcal{E}^R}similarity(e, e')$ for $e \in \mathcal{E}^L$.
Thus, we hypothesize that the distribution of maximum similarity values between exclusive nodes and those having a matching partner differ and can be used to distinguish those categories.
However, we note that the similarity distribution for already labeled nodes may differ from those that are not labeled, as the labeled nodes directly receive updates by a supervised loss signal.
Hence, we use \emph{historical} similarity values acquired when we selected unlabeled nodes for labeling, and the ground truth information about them having a match received after the labeling. 
Based on these, we fit two normal distributions for maximum similarities:
The first distribution with the probability function $P_{match}$ describes the distribution of maximal similarity score of nodes with matchings. Similarly, the function $P_{excl}$ computes the probability that the maximal similarity score belongs to an exclusive node. For each entity in question $e$, we take its maximal similarity score to the candidate in other graph  and compute a difference between two probabilities
$P_{match}(s_{max}(e)\leq x) - P_{excl}(s_{max}(e) \geq x)$ as heuristic score, c.f. Figure~\ref{fig:peb}.
This score is large if the maximal similarity of exclusive nodes is smaller than that of nodes with matchings.
We keep only entities with the score greater than threshold $t$, where $t$ is a hyperparameter. 
This way, we make sure that the score is used only if matching and exclusive nodes are distinguishable.
If there are not enough entities that fulfill this requirement, we use some simple fallback heuristic, e.g., degree, for the remaining nodes. 

\section{Evaluation Framework}
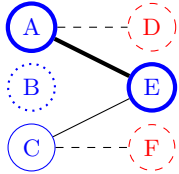
\begin{figure}
    \centering
    \begin{tikzpicture}[
        every node/.style={draw, circle},
        scale=.4,
        yscale=-1,
    ]
        \draw[blue, ultra thick] node (a) at (0, 0) {A};
        \draw[blue, thick, dotted] node (b) at (0, 2) {B};
        \draw[blue] node (c) at (0, 4) {C};
        \draw[red, dashed] node (d) at (4, 0) {D};
        \draw[blue, ultra thick] node (e) at (4, 2) {E};
        \draw[red, dashed] node (f) at (4, 4) {F};
        
        \draw (a) edge[dashed] (d);
        \draw (a) edge[ultra thick] (e);
        \draw (c) edge[dashed] (f);
        \draw (c) edge (e);
    \end{tikzpicture}
    \caption{
    Visualization of node categorisation for $\mathcal{E}^L = \{A, B, C\}$, and $\mathcal{E}^R = \{D, E, F\}$.
    Solid lines represent training alignments, whereas dashed ones denote test alignments.
    Node $B$ is the only exclusive node.
    All blue nodes are in the initial pool $\mathcal{P}_0$.
    The red dashed nodes $D$ and $F$ may not be requested for labeling as they neither are exclusive nor participate in a training alignment.
    When node $A$ is requested, only the alignment $(A, E)$ is returned, and $A$, as well as $E$, become unavailable.
    The second training alignment $(C, E)$ can still be obtained by requesting $C$.
    }
    \label{fig:node-categories}
\end{figure}
To evaluate active learning heuristics in-vitro, an alignment dataset comprising two graphs and labeled alignments is used.
These alignments are split into training alignments $\mathcal{A}_{train}$ and test alignments $\mathcal{A}_{test}$.
We employ an incremental batch-wise pool-based framework.
At step $i$, there is a pool of potential queries $\mathcal{P}_i \subseteq \left(\mathcal{E}^L \cup \mathcal{E}^R\right)$, from which a heuristic selects a fixed number of elements $\mathcal{Q}_i \subseteq \mathcal{P}_i$, where $b = |\mathcal{Q}_i|$ is often called the budget.
These queries are then passed to an alignment oracle $\mathfrak{O}$ simulating the labeling process and returning
$
\mathfrak{O}(\mathcal{Q}_i) = (\mathcal{A}_{i}, \mathcal{X}_i^L, \mathcal{X}_i^R)
$,
where the first component comprises the discovered alignments
$\mathcal{A}_{i} = \{(a, a') \in \mathcal{A}_{train} \mid \{a, a'\} \cap \mathcal{Q}_i \neq \emptyset\}$, and the last components the exclusive nodes $\mathcal{X}_i^L = \mathcal{X}^L \cap \mathcal{Q}_i$, and $\mathcal{X}_i^{R}$ analogously.
Afterward, the labeled nodes are removed from the pool, i.e. 
$
\mathcal{P}_{i+1} = \mathcal{P}_{i} \setminus \left(\mathcal{A}_i^L \cup \mathcal{A}_i^R \cup \mathcal{X}_i^L \cup \mathcal{X}_i^R\right)
$.
Note that when dealing with 1:n matchings, we remove all matches from the set of available nodes, despite some of them having additional alignment partners.
As each alignment edge can be retrieved using any of its endpoints, this does not pose a problem.
Now, the model is trained with all already found alignments, denoted by $\mathcal{A}_{\leq i}$, and without all exclusive nodes discovered so far, denoted by $\mathcal{X}_{\leq i}^L, \mathcal{X}_{\leq i}^R$, given as
$$
\mathcal{A}_{\leq i} = \bigcup_{j \leq i} \mathcal{A}_j,\quad
\mathcal{X}_{\leq i}^L = \bigcup_{j \leq i} \mathcal{X}_j^L,\quad
\mathcal{X}_{\leq i}^R = \bigcup_{j \leq i} \mathcal{X}_j^R.
$$
Following \cite{shen2017deep,ostapuk2019activelink}, we do not reset the parameters but warm-start the model with the previous iteration's parameters.
The pool is initialized with
$
\mathcal{P}_{0}:= \mathcal{A}_{train}^L \cup \mathcal{A}_{train}^R \cup \mathcal{X}^L \cup \mathcal{X}^R.
$
We exclude nodes that are not contained in the training alignment, but in the test alignments, as in this case, either a test alignment has to be revealed, or a node has to be unfaithfully classified as exclusive.
An illustration of the pool construction and an example query of size one is given in Figure~\ref{fig:node-categories}. \section{Experiments}
\subsection{Setup}
For evaluation, we use both subsets of the WK3l-15k dataset \cite{chen2017multigraph}\footnote{Note that the frequently used DBP15k dataset is not suitable for our experiments due to its construction. Exclusive nodes in DBP15K are exactly those having a degree of one and are therefore trivial to identify.}.
Similarly to \cite{pei2019semi} we extract additional entity alignments from the triple alignments.
Besides using the official train-test split, we perform an additional 80-20 train-validation split shared across all runs.
We additionally evaluate the transferability of the hyperparameter settings.
One of the challenges in active learning is that hyperparameter search for a new dataset is not possible because of the lack of labeled data at the beginning.
Therefore, for the evaluation of the second subset \texttt{en-fr}, we use the best hyperparameter settings which we obtained using \texttt{en-de} and compare how consistent are results for both subsets.

We employ a GNN-based model, GCN-Align \cite{wang2018cross}.
We use the best settings as found in \cite{berrendorf2019knowledge}.
To allow for Monte-Carlo Dropout estimation for the Bayesian heuristics, we additionally add a dropout layer between the embeddings and the GCN and vary the dropout rate.
We employ a margin-based matching loss, and we exclude so far identified exclusive nodes from the pool of negative samples.
Following \cite{beluch2018power}, we use 25 runs with different dropout masks for Bayesian approaches.
As evaluation protocol, we always retrieve 200 queries from the heuristic, update the exclusives and alignments using the oracle, and train the model for up to 4k epochs with early stopping on validation mean reciprocal rank (MRR) evaluated every 20 epochs, with a patience value of 200 epochs.
There are different scores for the evaluation of entity alignment, which evaluate different performance aspects \cite{berrendorf2020interpretable}.
In this work, we report Hits@1 (H@1) on the test alignments since this metric is most relevant for the applications.
We selected the heuristics' hyperparameters according to the AUC of the step vs. validation H@1 score.
Using the best hyperparameter configuration, we re-ran the experiments five times and report the mean and the standard deviation of the results on the test set.

\subsection{Results}
\textbf{Removal of exclusives --}
\begin{figure}\centering
    \includegraphics[width=.7\linewidth]{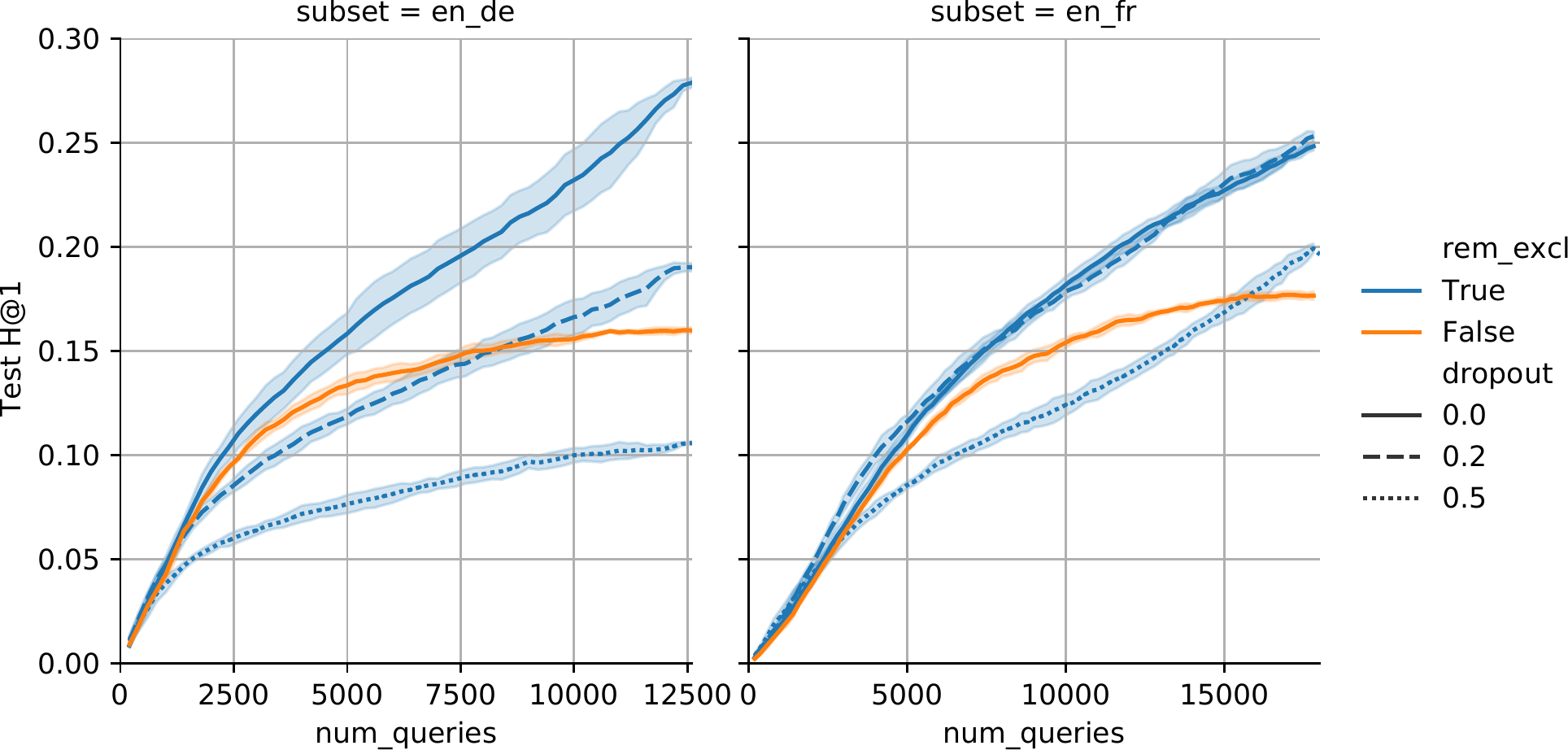}\\
    \caption{
    Performance vs number of queries for random baseline with different levels of dropout, and when removing exclusive nodes from message passing.
    Removing exclusives significantly improves the final performance.
    }
    \label{fig:excl_and_dropout}
\end{figure}
Figure~\ref{fig:excl_and_dropout} shows the test performance of the random selection baseline heuristic compared to the number of queries, with the standard deviation across five runs shown as shaded areas.
As can be seen by comparing the two solid lines, removing exclusives is advantageous, in particular, when many queries are performed, i.e., many exclusives are removed.
Therefore, we focus the subsequent analysis only on the case, when found exclusives are removed from the graph.
Moreover, we can see that using a high dropout value of 0.5 is disadvantageous on both datasets.
While a dropout value of 0.2 also hurts performance for the \texttt{en-de} subset, it does not have a negative influence on \texttt{en-fr}.
\begin{figure}\centering
    \includegraphics[width=.7\linewidth]{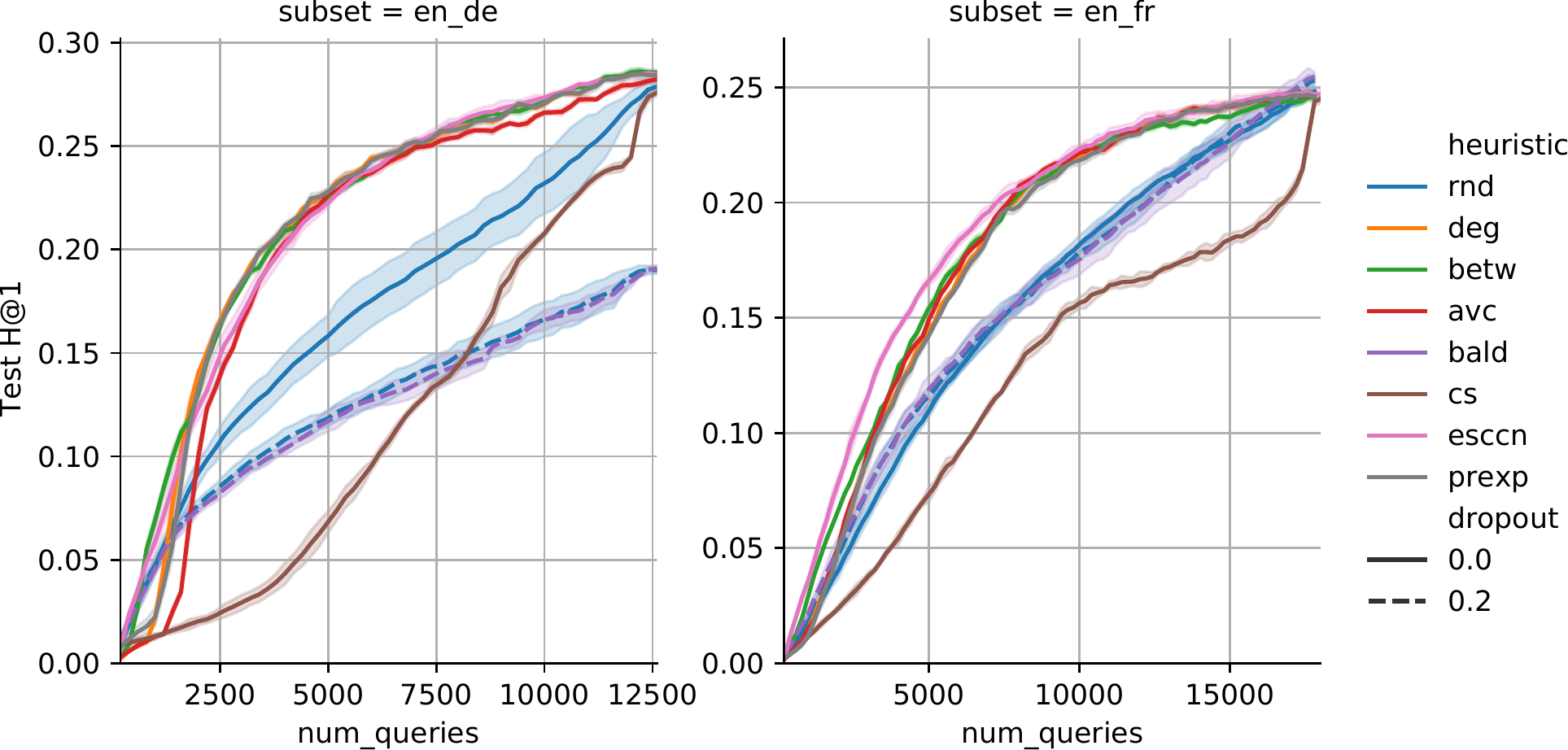}\\
    \caption{
    Performance on test alignments vs. number of queries for different heuristics.
    }
    \label{fig:step_vs_h1_all}
\end{figure}
\\
\textbf{Comparison of different heuristics --}
Figure~\ref{fig:step_vs_h1_all} compares the performance of different heuristics through all steps. Since there is a large overlap across different heuristics, we additionally compute AUC for each heuristic and report it in Table~\ref{tab:auc}.
\begin{table}\centering
    \caption{
    Mean and standard deviation of AUC number of queries vs. test hits @ 1 aggregated from five different runs for each heuristic and subset.
The * symbol indicates significant results compared to the \texttt{rnd} baseline according to unequal variances t-test (Welch's t-test) with $p < 0.01$.
}
    \begin{tabular*}{\linewidth}{l*{2}{@{\extracolsep{\fill}}l}}
    \toprule
    subset &                 en-de &                 en-fr \\
    \midrule
    avc       &  $0.2020 \pm 0.0005$* &  $0.1748 \pm 0.0005$* \\
    bald      &  $0.1222 \pm 0.0039$* &   $0.1514 \pm 0.0013$ \\
    betw      &  $0.2134 \pm 0.0005$* &  $0.1773 \pm 0.0004$* \\
    cs        &  $0.1117 \pm 0.0011$* &  $0.1185 \pm 0.0016$* \\
    deg       &  $0.2105 \pm 0.0005$* &  $0.1741 \pm 0.0005$* \\
    esccn     &  $0.2114 \pm 0.0006$* &  $0.1828 \pm 0.0021$* \\
    prexp     &  $0.2103 \pm 0.0009$* &  $0.1733 \pm 0.0009$* \\
    rnd       &   $0.1605 \pm 0.0040$ &   $0.1510 \pm 0.0019$ \\
    \bottomrule
    \end{tabular*}
    \label{tab:auc}
\end{table}
From the results, we observe that our expectations about the performance of different heuristics are mostly confirmed. Most of the heuristics perform significantly better than random sampling. Our intuitions about possible problems with \emph{coreset} in the context of entity alignment are also verified:
The heuristic performs consistently worse than the random sampling baseline. On the other hand, our new \emph{esccn} heuristic, which also tries to cover embedding space, but uses most central nodes instead, is one of the best performing heuristics.
We also observe an inferior performance of the uncertainty-based heuristic, which performance is comparable with the random heuristic. Note, that we also evaluated softmax entropy and maximal variation ratio heuristics from \cite{gao2018active} and their performance was similar. 
Overall, we see similar patterns for both subsets: There is a set of good performing heuristics and their performance is very similar.
\\
\textbf{Performance in earlier stages --}
\begin{figure}\centering
    \includegraphics[width=.7\linewidth]{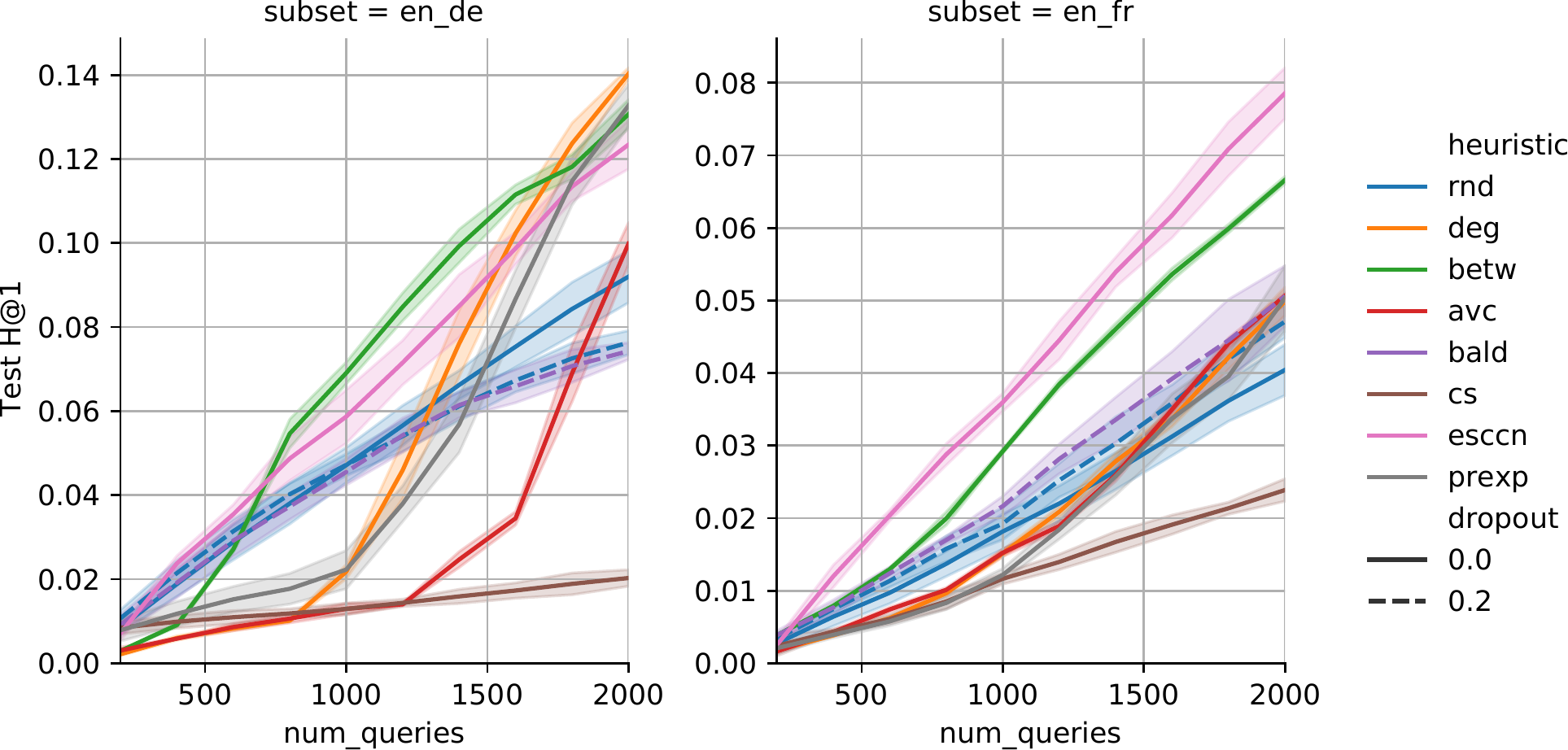}\\
    \caption{
    Performance on test alignments vs. number of queries for different heuristics.
    This figure shows only queries up to 2,000, i.e., the region where not many alignments have been found so far.
    }
    \label{fig:step_vs_h1}
\end{figure}
In many real-life applications, the labeling budget is limited; therefore, the model performance in the first steps is of higher relevance.
Therefore, in Figure~\ref{fig:step_vs_h1}, we analyze the model performance in the first 2,000 iterations.
We observe that the \emph{escnn} and \emph{betw} heuristics compete for first prize and that towards the end, they are superseded by other heuristics. 
\begin{figure}\centering
    \includegraphics[width=.7\linewidth]{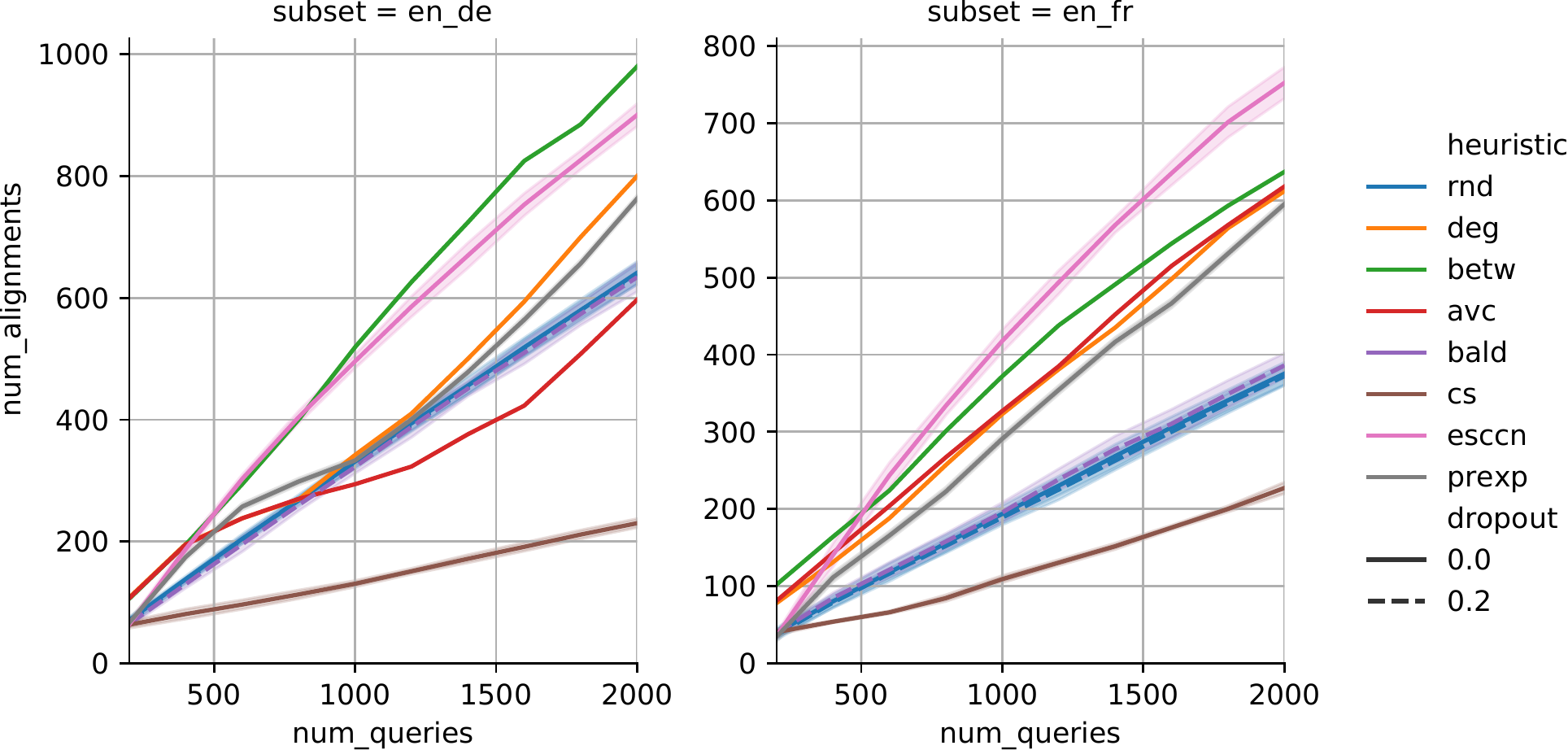}\\
    \caption{
    Number of found training alignments vs. number of queries for different heuristics.
    This figure shows only queries up to 2,000, i.e., the region where not many alignments have been found so far.
    }
    \label{fig:step_vs_pairs}
\end{figure}
\\
\textbf{Influence of positive matchings --} 
In Figure~\ref{fig:step_vs_pairs}, we show the number of alignment pairs identified by each heuristic in the first 2,000 steps. 
For most heuristics, the plots look very similar to the plots in Figure~\ref{fig:step_vs_h1} above with the performance on the $y$ axis. 
In Figure~\ref{fig:excl_and_dropout}, we also saw that the removal of exclusive nodes affects the performance only at later iterations. 
Therefore, we can conclude that finding nodes with matches is especially important in the early training stages.

On the whole, we can conclude that node centrality based heuristics like \emph{betw} are the right choice for active learning for entity alignment. It achieves performance comparable with model-based approaches and does not require access to model predictions during the labeling process. The labeling ordering can be precomputed and does not change, also facilitating to parallelize the labeling process for a fixed budget to multiple annotators, e.g., using systems such as Amazon Mechanical Turk.
 \section{Conclusion}
In this paper, we introduced the novel task of active learning for entity alignment and discussed its differences to the classical active learning setting.
Moreover, we proposed several different heuristics, both, adaptions of existing heuristics used for classification, as well as heuristics specifically designed for this particular task.
In a thorough empirical analysis, we showed strong performance of simple centrality and graph cover heuristics, while adaptations of state-of-the-art heuristics for classification showed inferior performance.
For future work, we envision transferring our approaches to other graph matching problems, such as matching road networks \cite{faerman2019graph} or approximating graph edit distance \cite{DBLP:conf/icml/LiGDVK19}.
Moreover, we aim to study the generalization of our findings to other datasets and models. 
\section*{Acknowledgements}
This work has been funded by the German Federal Ministry of Education and Research (BMBF) under Grant No. 01IS18036A. The authors of this work take full responsibilities for its content.

\bibliographystyle{splncs04}
\bibliography{main}

\begin{thebibliography}{10}
\providecommand{\url}[1]{\texttt{#1}}
\providecommand{\urlprefix}{URL }
\providecommand{\doi}[1]{https://doi.org/#1}

\bibitem{bast2016semantic}
Bast, H., Bj{\"o}rn, B., Haussmann, E.: Semantic search on text and knowledge
  bases. Foundations and Trends in Information Retrieval  \textbf{10}(2-3),
  119--271 (2016)

\bibitem{beluch2018power}
Beluch, W.H., Genewein, T., N{\"{u}}rnberger, A., K{\"{o}}hler, J.M.: The power
  of ensembles for active learning in image classification. In: {CVPR}. pp.
  9368--9377. {IEEE} Computer Society (2018)

\bibitem{berrendorf2019knowledge}
Berrendorf, M., Faerman, E., Melnychuk, V., Tresp, V., Seidl, T.: Knowledge
  graph entity alignment with graph convolutional networks: Lessons learned.
  arXiv preprint arXiv:1911.08342  (2019)

\bibitem{berrendorf2020interpretable}
Berrendorf, M., Faerman, E., Vermue, L., Tresp, V.: Interpretable and fair
  comparison of link prediction or entity alignment methods with adjusted mean
  rank. arXiv preprint arXiv:2002.06914  (2020)

\bibitem{cai2017active}
Cai, H., Zheng, V.W., Chang, K.C.C.: Active learning for graph embedding. arXiv
  preprint arXiv:1705.05085  (2017)

\bibitem{cao2019multi}
Cao, Y., Liu, Z., Li, C., Liu, Z., Li, J., Chua, T.: Multi-channel graph neural
  network for entity alignment. In: {ACL} {(1)}. pp. 1452--1461. ACL (2019)

\bibitem{chen2017multigraph}
Chen, M., Tian, Y., Yang, M., Zaniolo, C.: Multilingual knowledge graph
  embeddings for cross-lingual knowledge alignment. In: {IJCAI}. pp.
  1511--1517. ijcai.org (2017)

\bibitem{cortes2013active}
Cort{\'{e}}s, X., Serratosa, F.: Active-learning query strategies applied to
  select a graph node given a graph labelling. In: GbRPR. Lecture Notes in
  Computer Science, vol.~7877, pp. 61--70. Springer (2013)

\bibitem{cortes2012active}
Cort{\'{e}}s, X., Serratosa, F., Sol{\'{e}}{-}Ribalta, A.: Active graph
  matching based on pairwise probabilities between nodes. In: {SSPR/SPR}.
  Lecture Notes in Computer Science, vol.~7626, pp. 98--106. Springer (2012)

\bibitem{das2018study}
Das, K., Samanta, S., Pal, M.: Study on centrality measures in social networks:
  a survey. Social Netw. Analys. Mining  \textbf{8}(1), ~13 (2018)

\bibitem{irtutorial}
Dietz, L., Kotov, A., Meij, E.: Utilizing knowledge graphs for text-centric
  information retrieval. In: The 41st International ACM SIGIR Conference on
  Research \& Development in Information Retrieval. p. 1387–1390. SIGIR '18,
  Association for Computing Machinery, New York, NY, USA (2018).
  \doi{10.1145/3209978.3210187}, \url{https://doi.org/10.1145/3209978.3210187}

\bibitem{faerman2018lasagne}
Faerman, E., Borutta, F., Fountoulakis, K., Mahoney, M.W.: {LASAGNE:} locality
  and structure aware graph node embedding. In: {WI}. pp. 246--253. {IEEE}
  Computer Society (2018)

\bibitem{faerman2019graph}
Faerman, E., Voggenreiter, O., Borutta, F., Emrich, T., Berrendorf, M.,
  Schubert, M.: Graph alignment networks with node matching scores. In: Graph
  Representation Learning NeurIPS 2019 Workshop (2019)

\bibitem{gal2016uncertainty}
Gal, Y.: Uncertainty in deep learning. Ph.D. thesis, PhD thesis, University of
  Cambridge (2016)

\bibitem{gal2016dropout}
Gal, Y., Ghahramani, Z.: Dropout as a bayesian approximation: Representing
  model uncertainty in deep learning. In: {ICML}. {JMLR} Workshop and
  Conference Proceedings, vol.~48, pp. 1050--1059. JMLR.org (2016)

\bibitem{gal2017deep}
Gal, Y., Islam, R., Ghahramani, Z.: Deep bayesian active learning with image
  data. In: {ICML}. Proceedings of Machine Learning Research, vol.~70, pp.
  1183--1192. {PMLR} (2017)

\bibitem{gao2018active}
Gao, L., Yang, H., Zhou, C., Wu, J., Pan, S., Hu, Y.: Active discriminative
  network representation learning. In: {IJCAI}. pp. 2142--2148. ijcai.org
  (2018)

\bibitem{geifman2017deep}
Geifman, Y., El-Yaniv, R.: Deep active learning over the long tail. arXiv
  preprint arXiv:1711.00941  (2017)

\bibitem{guo2019learning}
Guo, L., Sun, Z., Hu, W.: Learning to exploit long-term relational dependencies
  in knowledge graphs. In: {ICML}. Proceedings of Machine Learning Research,
  vol.~97, pp. 2505--2514. {PMLR} (2019)

\bibitem{hinton2015distilling}
Hinton, G., Vinyals, O., Dean, J.: Distilling the knowledge in a neural
  network. arXiv preprint arXiv:1503.02531  (2015)

\bibitem{houlsby2011bayesian}
Houlsby, N., Husz{\'a}r, F., Ghahramani, Z., Lengyel, M.: Bayesian active
  learning for classification and preference learning. arXiv preprint
  arXiv:1112.5745  (2011)

\bibitem{lewis1994heterogeneous}
Lewis, D.D., Catlett, J.: Heterogeneous uncertainty sampling for supervised
  learning. In: {ICML}. pp. 148--156. Morgan Kaufmann (1994)

\bibitem{li2019semi}
Li, C., Cao, Y., Hou, L., Shi, J., Li, J., Chua, T.: Semi-supervised entity
  alignment via joint knowledge embedding model and cross-graph model. In:
  {EMNLP/IJCNLP} {(1)}. pp. 2723--2732. ACL (2019)

\bibitem{DBLP:conf/icml/LiGDVK19}
Li, Y., Gu, C., Dullien, T., Vinyals, O., Kohli, P.: Graph matching networks
  for learning the similarity of graph structured objects. In: {ICML}.
  Proceedings of Machine Learning Research, vol.~97, pp. 3835--3845. {PMLR}
  (2019)

\bibitem{DBLP:conf/cidr/MahdisoltaniBS15}
Mahdisoltani, F., Biega, J., Suchanek, F.M.: {YAGO3:} {A} knowledge base from
  multilingual wikipedias. In: {CIDR}. www.cidrdb.org (2015)

\bibitem{malmi2017active}
Malmi, E., Gionis, A., Terzi, E.: Active network alignment: a matching-based
  approach. In: Proceedings of the 2017 ACM on Conference on Information and
  Knowledge Management. pp. 1687--1696 (2017)

\bibitem{ostapuk2019activelink}
Ostapuk, N., Yang, J., Cudr{\'{e}}{-}Mauroux, P.: Activelink: Deep active
  learning for link prediction in knowledge graphs. In: {WWW}. pp. 1398--1408.
  {ACM} (2019)

\bibitem{pei2019semi}
Pei, S., Yu, L., Hoehndorf, R., Zhang, X.: Semi-supervised entity alignment via
  knowledge graph embedding with awareness of degree difference. In: {WWW}. pp.
  3130--3136. {ACM} (2019)

\bibitem{Puthal2015}
Puthal, D., Nepal, S., Paris, C., Ranjan, R., Chen, J.: Efficient algorithms
  for social network coverage and reach. In: BigData Congress. pp. 467--474.
  {IEEE} Computer Society (2015)

\bibitem{sener2017active}
Sener, O., Savarese, S.: Active learning for convolutional neural networks: {A}
  core-set approach. In: {ICLR} (Poster). OpenReview.net (2018)

\bibitem{settles2009active}
Settles, B.: Active learning literature survey. Tech. rep., University of
  Wisconsin-Madison Department of Computer Sciences (2009)

\bibitem{shen2017deep}
Shen, Y., Yun, H., Lipton, Z.C., Kronrod, Y., Anandkumar, A.: Deep active
  learning for named entity recognition. In: {ICLR} (Poster). OpenReview.net
  (2018)

\bibitem{speer2017conceptnet}
Speer, R., Chin, J., Havasi, C.: Conceptnet 5.5: An open multilingual graph of
  general knowledge. In: {AAAI}. pp. 4444--4451. {AAAI} Press (2017)

\bibitem{sun2017cross}
Sun, Z., Hu, W., Li, C.: Cross-lingual entity alignment via joint
  attribute-preserving embedding. In: ISWC {(1)}. Lecture Notes in Computer
  Science, vol. 10587, pp. 628--644. Springer (2017)

\bibitem{sun2018bootstrapping}
Sun, Z., Hu, W., Zhang, Q., Qu, Y.: Bootstrapping entity alignment with
  knowledge graph embedding. In: IJCAI. pp. 4396--4402 (2018)

\bibitem{sun2019knowledge}
Sun, Z., Wang, C., Hu, W., Chen, M., Dai, J., Zhang, W., Qu, Y.: Knowledge
  graph alignment network with gated multi-hop neighborhood aggregation. arXiv
  preprint arXiv:1911.08936  (2019)

\bibitem{trisedya2019entity}
Trisedya, B.D., Qi, J., Zhang, R.: Entity alignment between knowledge graphs
  using attribute embeddings. In: {AAAI}. pp. 297--304. {AAAI} Press (2019)

\bibitem{vrandevcic2014wikidata}
Vrande{\v{c}}i{\'c}, D., Kr{\"o}tzsch, M.: Wikidata: a free collaborative
  knowledgebase. Communications of the ACM  \textbf{57}(10),  78--85 (2014)

\bibitem{wang2016cost}
Wang, K., Zhang, D., Li, Y., Zhang, R., Lin, L.: Cost-effective active learning
  for deep image classification. {IEEE} Trans. Circuits Syst. Video Techn.
  \textbf{27}(12),  2591--2600 (2017)

\bibitem{wang2018cross}
Wang, Z., Lv, Q., Lan, X., Zhang, Y.: Cross-lingual knowledge graph alignment
  via graph convolutional networks. In: {EMNLP}. pp. 349--357. ACL (2018)

\bibitem{wu2019active}
Wu, Y., Xu, Y., Singh, A., Yang, Y., Dubrawski, A.: Active learning for graph
  neural networks via node feature propagation. arXiv preprint arXiv:1910.07567
   (2019)

\bibitem{xu2019crosslingual}
Xu, K., Wang, L., Yu, M., Feng, Y., Song, Y., Wang, Z., Yu, D.: Cross-lingual
  knowledge graph alignment via graph matching neural network. In: {ACL} {(1)}.
  pp. 3156--3161. ACL (2019)

\bibitem{yang2017suggestive}
Yang, L., Zhang, Y., Chen, J., Zhang, S., Chen, D.Z.: Suggestive annotation:
  {A} deep active learning framework for biomedical image segmentation. In:
  {MICCAI} {(3)}. Lecture Notes in Computer Science, vol. 10435, pp. 399--407.
  Springer (2017)

\bibitem{zhang2019multi}
Zhang, Q., Sun, Z., Hu, W., Chen, M., Guo, L., Qu, Y.: Multi-view knowledge
  graph embedding for entity alignment. In: {IJCAI}. pp. 5429--5435. ijcai.org
  (2019)

\bibitem{zhang2017active}
Zhang, Y., Lease, M., Wallace, B.C.: Active discriminative text representation
  learning. In: {AAAI}. pp. 3386--3392. {AAAI} Press (2017)

\bibitem{zhu2019neighborhood}
Zhu, Q., Zhou, X., Wu, J., Tan, J., Guo, L.: Neighborhood-aware attentional
  representation for multilingual knowledge graphs. In: {IJCAI}. pp.
  1943--1949. ijcai.org (2019)

\end{thebibliography}

\end{document}